\documentclass[runningheads]{llncs}

\usepackage[T1]{fontenc}

\usepackage{graphicx,verbatim}
\usepackage{amsmath,amssymb,amsfonts}
\usepackage{multirow}
\usepackage{threeparttable}
\usepackage{cite}
\usepackage{hyperref}
\usepackage{CJKutf8}

\begin{document}

\makeatletter
\renewcommand{\@fnsymbol}[1]{\ifcase#1 \or *\or *\fi}

\title{GIGP: A Global Information Interacting and Geometric Priors Focusing Framework for Semi-supervised Medical Image Segmentation}

\author{
Lianyuan Yu\inst{1} \and 
Xiuzhen Guo\inst{1} \and 
Ji Shi\inst{2}\thanks{Corresponding author}  \and  
Hongxiao Wang\inst{2} \and 
Hongwei Li\inst{1}\textsuperscript{*}   
}
\authorrunning{Lianyuan Yu et al.}
\institute{The School of Mathematical Science, Capital Normal University, Beijing 100048, China \and
The Academy for Multidisciplinary Studies, Capital Normal University, Beijing 100048, China \\
\email{lianyuanyu8@gmail.com; xiuzhenguo589@gmail.com; shiji@cnu.edu.cn; hwang21@cnu.edu.cn; hongwei.li91@cnu.edu.cn}}

\maketitle             

\begin{abstract}
Semi-supervised learning enhances medical image segmentation by leveraging unlabeled data, reducing reliance on extensive labeled datasets. On the one hand, the distribution discrepancy between limited labeled data and abundant unlabeled data can hinder model generalization. Most existing methods rely on local similarity matching, which may introduce bias. In contrast, Mamba effectively models global context with linear complexity, learning more comprehensive data representations. On the other hand, medical images usually exhibit consistent anatomical structures defined by geometric features. Most existing methods fail to fully utilize global geometric priors, such as volumes, moments etc. In this work, we introduce a global information interaction and geometric priors focus framework (GIGP). Firstly, we present a Global Information Interaction Mamba module to reduce distribution discrepancy between labeled and unlabeled data. Secondly, we propose a Geometric Moment Attention Mechanism to extract richer global geometric features. Finally, we propose Global Geometric Perturbation Consistency to simulate organ dynamics and geometric variations, enhancing the ability of the model to learn generalized features. The superior performance on the NIH Pancreas and Left Atrium datasets demonstrates the effectiveness of our approach.


\end{abstract}

\section{Introduction}
Medical image segmentation is vital for computer-aided diagnosis, yet deep learning algorithms often require extensive labeled data. The intricate nature of medical images, including complex backgrounds and blurred edges, complicates labeling, restricting dataset availability. Semi-supervised learning bridges this gap by leveraging both labeled and unlabeled data, enhancing model performance and advancing medical imaging training efficiency.

Semi-supervised medical image segmentation methods include consistency regularization \cite{b3,b4,b5,b6,b7,b8}, pseudo-labeling \cite{b52}, adversarial learning \cite{b57}, and graph-based approaches \cite{b58}. Consistency regularization is based on the premise that predictions of a model for a sample should ideally be robust to minor changes. To enhance information extraction from unlabeled data, methods like uncertainty mapping \cite{b5}, multi-scale consistency \cite{b6}, and similarity matrices between labeled and unlabeled features \cite{b7} are used. However, due to the complex backgrounds inherent in medical images, the information derived from unlabeled samples often contains noise, limiting the effectiveness of general models.

In medical data, the same organ typically exhibits consistent anatomical structures, which are closely related to geometric features, as anatomical structures are inherently defined by specific geometric attributes such as shape, size, position, and orientation. Some studies learn the shape of target objects by utilizing signed distance maps \cite{b3, b4} or constructing shape dictionaries \cite{b20}. Other research proposes methods such as volume constancy constraints \cite{b8}, learning skeleton and boundary structures \cite{b21}, and learning polygon vertices of samples \cite{b22}. These geometric feature extraction methods effectively improve segmentation performance but often rely on a limited variety of geometric features. Meanwhile, due to the inherent limitations of CNNs, the receptive field is constrained by the size of convolutional kernels, primarily capturing local patterns. Therefore, incorporating more types of global geometric features is crucial for enabling the model to learn more discriminative and semantically meaningful representations from abundant unlabeled data.

Medical datasets usually have a small labeled subset and a large unlabeled one, with the labeled data insufficiently representing the full diversity, leading to distribution gaps. Bai et al. \cite{b60} introduced a copy-paste method to propagate semantic information from labeled to unlabeled data. Other approaches assume that semantically similar regions between labeled and unlabeled data, drawn from the same distribution, can share labels \cite{b13, b7, b14}. Techniques such as cosine similarity \cite{b13}, omni-correlation matrices \cite{b7}, and graph-based methods \cite{b14} have been used to identify and transfer label information. However, these methods mainly focus on local similarity matching, so they may not effectively capture the overall data distribution, which is essential for mitigating the bias between the scarce labeled and the ample unlabeled data.

Inspired by the above, we propose a novel semi-supervised medical image segmentation framework (GIGP) from a global perspective. Firstly, Mamba can integrate global contextual information with linear computational complexity, which enables it to better align the shared features between labeled and unlabeled data. Secondly, geometric moments can capture various global geometric features of samples, such as shape, symmetry, and directionality, helping characterize the anatomical structures of samples. Meanwhile, normalized geometric moments provide scale invariance and preserve information during multi-scale fusion, reducing information loss from upsampling or downsampling. Thirdly, dynamic changes and geometric variations naturally occur in the same organ across individuals or over time. Sine wave transformations effectively simulate these variations, such as the periodic shape changes of the atrium during heartbeats and the morphological differences in the pancreas among individuals.

In summary, our contributions are fourfold: (1) We propose a Global Information Interaction Mamba (GIIM) module that jointly learns features from labeled and unlabeled data at corresponding spatial locations. By transferring parameters learned from labeled data to unlabeled data globally, GIIM captures shared characteristics more effectively, reducing their distribution gap and improving overall learning performance. (2) We propose a Geometric Moment Attention Mechanism (GMAM) that enforces multi-view and multi-scale constraints using geometric moments, offering comprehensive guidance on diverse geometric features. The attention mechanism weights these features, allowing the model to precisely focus on and extract key geometric information from images. (3) We propose a Global Geometric Perturbation Consistency (GGPC) that uses periodic sine wave distortions to simulate organ dynamics and geometric variations, improving the adaptability of the model to anatomical variability in medical imaging. (4) Experiments on the NIH Pancreas \cite{b16} and Left Atrium \cite{b17} datasets demonstrate that our method significantly enhances model generalization.

\begin{figure}[!t]
\center
\includegraphics[width=\textwidth]{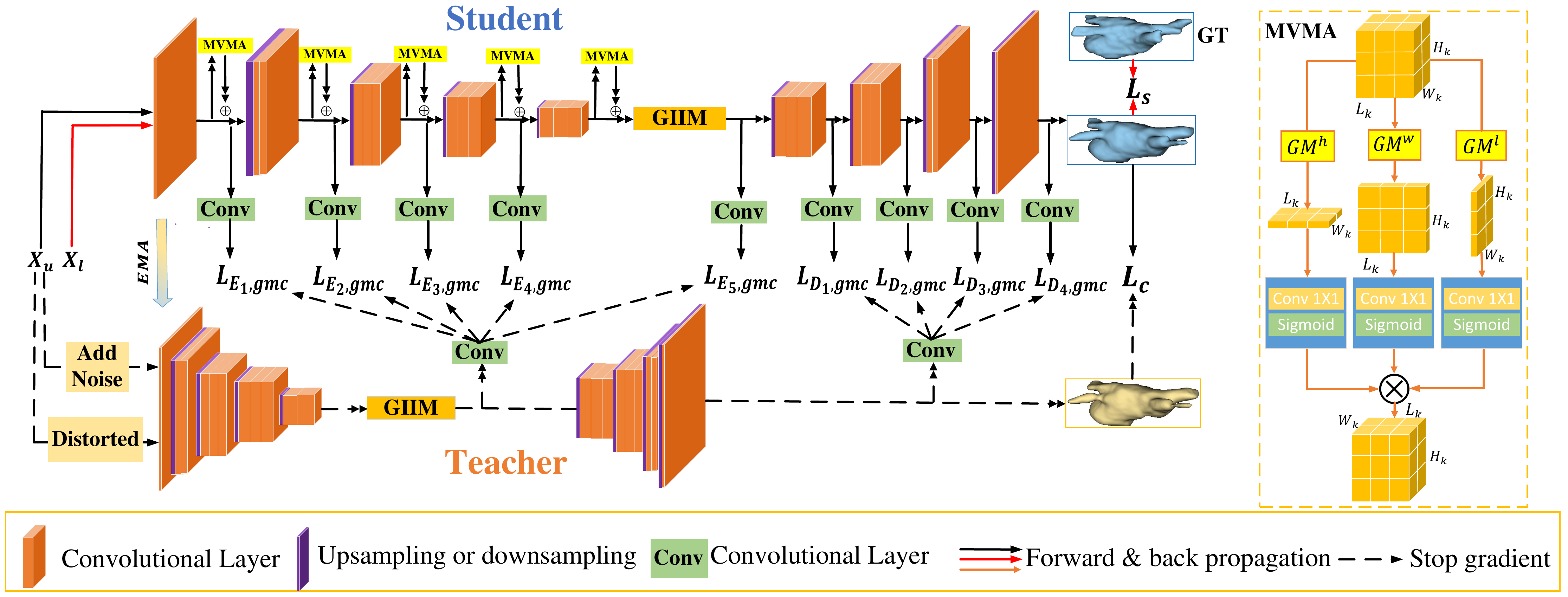}
\caption{The overview of GIGP. $L_{E_k,gmc}$ and $L_{D_k,gmc}$ represent the the moment losses of the k-th layer of the encoder and decoder, respectively. $GM^{h}$, $GM^{w}$, and $GM^{l}$ represent computing geometric moments in three dimensions, respectively.}
\label{fig1}
\end{figure}

\section{Methodology}
\label{section 2}

\subsection{Overview}
\label{section 2.1}

Our framework, illustrated in Fig. \ref{fig1}, uses the Mean Teacher (MT) as the backbone. In each forward pass, a mini-batch includes $m$ labeled samples $\left\{X_{l_i}\right\}_{i=1}^m$ and $n$ unlabeled samples $\left\{X_{u_j}\right\}_{j=1}^n$. The student network processes both original labeled and unlabeled samples, while the teacher network processes noise version and global geometric perturbation version of unlabeled samples. The MT network then generates probability maps $P^s_{l}=\left\{P^s_{l_i}\right\}_{i=1}^m$, $P^s_{u}=\left\{P^s_{u_j}\right\}_{j=1}^n$, $P^{t,ns}_{u}=\left\{P^{t,ns}_{u_j}\right\}_{j=1}^n$, and $P^{t,gp}_{u}=\left\{P^{t,gp}_{u_j}\right\}_{j=1}^n$. Labels $Y_{l}=\left\{Y_{l_i}\right\}_{i=1}^m$ supervise input labeled data. The total loss of GIGP can be formulated as:
\begin{equation}
L=L_s+\gamma_1L_c+\gamma_2L_{gmc},
\end{equation}
where $\gamma_1,\gamma_2$ are the weight coefficients and $L_{gmc}$ is introduced in Sec. \ref{section 2.3}. $
L_s=D(P^s_l, Y_l)+\alpha_E{E(P^s_l, Y_l)}$ and $L_c=MSE(P^s_u, P^{t,ns}_u)+MSE(P^s_u,P^{t,gp}_u)$.
The dice coefficient $D(\cdot)$ and cross-entropy $E(\cdot)$ guide the training for labeled data. $\alpha_E$ is the weight coefficient. $MSE(\cdot)$ is the mean square error function. 

\subsection{Global Information Interaction Mamba}
\label{section 2.2}

\begin{figure}[!t]
\center
\includegraphics[width=\textwidth]{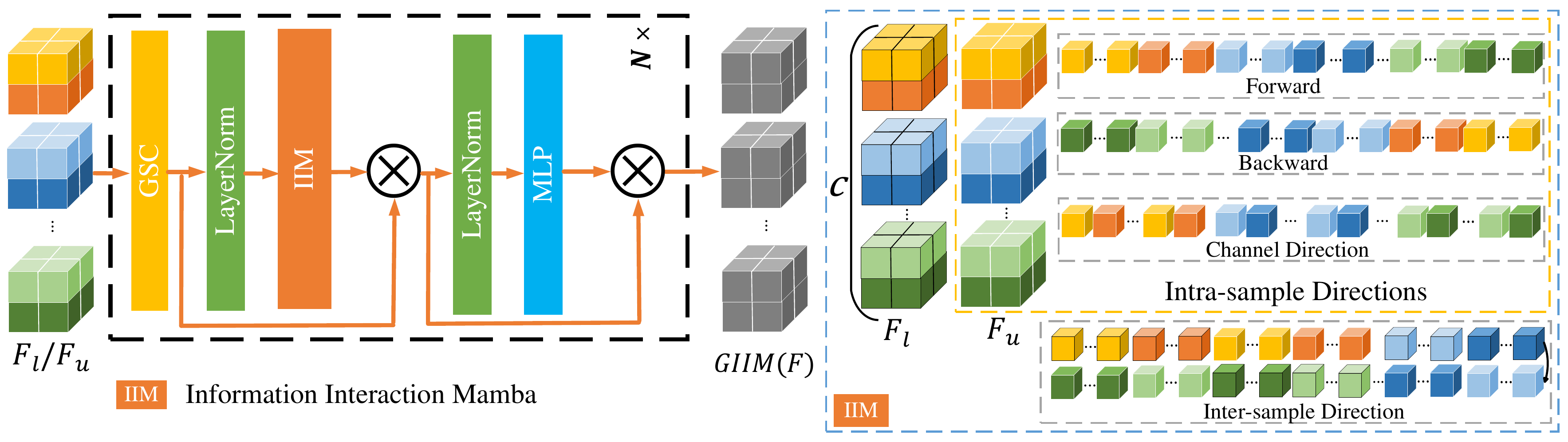} 
\caption{The overview of Geometric Information Propagation Mamba. GSC is the gated spatial convolution module \cite{b30}. LayerNorm denotes the layer normalization, and MLP represents the multiple layers perception layer to enrich the feature representation. IIM is the Information Interaction Mamba.}
\label{fig3}
\end{figure}

GIIM module leverages Mamba \cite{b43} theory to enable channel-wise interaction between labeled and unlabeled features, effectively encoding and learning from their shared representations.

As illustrated in Fig. \ref{fig3}, IIM comprises four directional components: forward direction $f$, reverse direction $r$, inter-channel direction $c$, and inter-sample direction $t$. The output of IIM is defined as:
\begin{equation}
IIM(F)=Mamba(F_f)+Mamba(F_r)+\lambda_{1}Mamba(F_c)
+\lambda_{2}Mamba(F_t),
\end{equation}
where $\lambda_{1}$ and $\lambda_{2}$ are weighting coefficients, and $Mamba$ denotes the Mamba layer, which models global information within a sequence.

\subsection{Geometric Moment Attention Mechanism}
\label{section 2.3}
GMAM constructs a multi-view second-order geometric moment attention mechanism (MVMA) and a multi-scale second-order normalized geometric moment consistency (MSGC), as shown in Fig. \ref{fig1}. Second-order geometric moments assist the model in capturing the symmetry, shape, and orientation of samples. MVMA is capable of extracting global geometric features from the three dimensions of 3D features, which helps the model to understand the three-dimensional spatial structure comprehensively. MSGC utilizes scale invariance to achieve the complementation of deep and shallow geometric features.

MVMA is applied after each encoder layer. The k-th encoder layer outputs $P_{k}\in{R^{{C_k}\times{H_k}\times{W_k}\times{L_k}}}$. MVMA computes second-order geometric moments along different directions: 
\begin{equation}
\mu^h_{k,pqr}=\sum_{x}(x-x_0)^p(y-y_0)^q(z-z_0)^rP_k(x,y,z) , 
\end{equation}
where $(x_0, y_0, z_0)$ is the centroid of images, $p,q,r \in {\mathbb{N}},p+q+r=2$. $\mu^h_{k,pqr}\in{{\mathbb{R}}^{{C_k}\times{1}\times{W_k}\times{L_k}}}$, the same applies to the $y$ and $z$ directions. $\mu^h_{k,pqr}$, $\mu^w_{k,pqr}$, and $\mu^l_{k,pqr}$ include global geometric information of different dimensions. Then we construct their spatial attention map:
\begin{equation}
z_k(x, y, z) = P_k(x, y, z) + \lambda_P \cdot P_k(x, y, z) \cdot \prod_{i \in \{h,w,r\}} \sigma\left(\text{Conv}(\mu^i_{k,pqr})\right),
\end{equation}
where $\lambda_{P}$ is the coefficient learned by the neural network, $\sigma(\cdot)$ is the sigmoid function, and \text{Conv} is $1\times{1}$ convolutions. MVMA includes the DiVA module proposed by Behjati et al.\cite{b36} as a special case in two-dimensional images.

MSGC is applied in each resolution layer in the student network, as well as in the final layers of both the encoder and decoder in the teacher network. Their normalized geometric moments can be formulated as:  
\begin{equation}
\mu^s_{E_k/D_k,pqr}=\sum_{x,y,z}((x-x_0)/\sigma_x)^p ((y-y_0)/\sigma_y)^q 
((z-z_0)/\sigma_z)^rf^s_{E_k/D_k}(x,y,z),
\end{equation}
\begin{equation}
\mu^{t,ns/gp}_{E/D,pqr}=\sum_{x,y,z}((x-x_0)/\sigma_x)^p((y-y_0)/\sigma_y)^q
((z-z_0)/\sigma_z)^rf^{t,ns/gp}_{E/D}(x,y,z),
\end{equation}
where $\sigma_x=\sqrt{(1/{H_k})\sum_{x=1}^{H_k}(x-x_0)^2}$, the same applies to the $y$ and $z$ directions. $f^s_{E_k/D_k}$ and $f^{t,ns/gp}_{E/D}$ are features of the encoder or decoder after being transformed into a single channel for channel alignment. Our MSGC loss is defined as follows:
\begin{equation}
L_{gmc}=\sum_{k=1}^K\sum_{p,q,r}(\alpha_k | \mu^s_{E_k/D_k,pqr}-\mu^{t,ns}_{E/D,pqr} |+\beta_k | \mu^s_{E_k/D_k,pqr}-\mu^{t,gp}_{E/D,pqr} |),
\end{equation}
where $K$ represents the number of different resolution layers in the encoder or decoder. $\alpha_k$ and $\beta_k$ are weight coefficients. 

\subsection{Global Geometric Perturbation Consistency}
\label{section 2.4}

GGPC introduces a periodic distortion effect by superimposing a sine wave perturbation on each axis of the three-dimensional grid, thereby deforming the original coordinates to generate a distorted grid with wave-like characteristics.

Firstly, we generate a wave grid to distort the three-dimensional coordinates using the following formula:
\begin{align}
\textbf{G}(x, y, z) = 
\begin{bmatrix}
x + A \sin(2 \pi f x) \\
y + A \sin(2 \pi f y) \\
z + A \sin(2 \pi f z)
\end{bmatrix},
\end{align}
where $\textbf{G}$ represents distorted grid coordinates, $A$ is the amplitude and $f$ is the frequency. They make the distortions sufficiently subtle, enabling the model to capture the subtle deformations in the input data.

Secondly, utilize three-dimensional interpolation techniques to map $n$ unlabeled samples $\left\{X_{u_j}\right\}_{j=1}^n$ to the distorted grid, denoted as:
\begin{equation}
O_{u_j} = \textbf{F}_{\text{grid}}(X_{u_j},\textbf{G})
\end{equation}
where $\textbf{F}_{\text{grid}}$ is the function used to perform interpolation. 

\section{Experiments}
\label{section 5}

\textbf{Datasets.}
Our method is evaluated on NIH pancreas (Pancreas) and Left Atrium (LA) datasets. The Pancreas Dataset \cite{b16} consists of 82 contrast$-$enhanced 3D abdominal CT scans with manual annotations. Each CT volume has a size ranging from $512\times{512}\times{181}$ to $512\times{512}\times{466}$. In the experiments, we apply a soft tissue CT window of $[{-}120, 240]$ HU. The CT scans are cropped around the pancreas region, with the margins extended by 25 voxels. A total of 62 volumes are used for training, and 20 volumes are used for testing. The 3D Left Atrium (LA) segmentation dataset \cite{b17} is from the 2018 Left Atrium Segmentation Challenge consisting of 100 3D gadolinium-enhanced MR images, each with a resolution of ${0.625^3mm^3}$. In our experiments, 80 scans are used for training and 20 scans for testing. All images are cropped to the center of the heart region and normalized to have zero mean and unit variance.

\begin{table}[!t]
\centering
\caption{Results in quantitative comparison on Pancreas dataset.}
\renewcommand{\arraystretch}{1.2}
\begin{threeparttable}
\fontsize{8pt}{10pt}\selectfont
\begin{tabular}{@{}cccccc@{}}
\hline
\multirow{2}{*}{Method} & Scans used & \multicolumn{4}{c}{Metrics}\\
\cline{2-6}
& Labeled/Unlabeled & Dice($\%$) $\uparrow$ & Jaccard($\%$) $\uparrow$ & 95HD $\downarrow$ & ASD $\downarrow$\\
\hline
V-Net \cite{b42} & [6/0,12/0] & [60.39,71.52] & [46.17,57.68] & [24.94,18.12] & [2.23,5.41]\\ 
\hline
DTC \cite{b3} & [6/56,12/50] & [69.01,73.55] & [54.52,59.90] & [20.99,13.55] & [2.33,1.59]\\ 
SASSnet \cite{b4} & [6/56,12/50] & [72.28,77.43] & [58.06,64.18] & [14.30,11.78] & [2.45,1.53]\\
UA-MT \cite{b38} & [6/56,12/50] & [71.26,76.66] & [56.15,63.09] & [22.01,11.85] & [7.36,3.53]\\
URPC \cite{b6} & [6/56,12/50] & [72.66,75.22] & [18.99,22.75] & [22.63,13.86] & [6.36,4.06]\\
MC-Net \cite{b39} & [6/56,12/50] & [69.96,74.80] & [55.57,60.57] & [23.90,19.18] & [\textbf{1.56},4.64]\\
MC-Net+ \cite{b40} & [6/56,12/50] & [63.31,73.71] & [49.15,60.34] & [31.99,13.93] & [2.09,4.00]\\
AC-MT \cite{b5} & [6/56,12/50] & [73.00,78.86] & [58.93,66.02] & [18.36,7.98] & [1.91,1.47]\\
CAML \cite{b7} & [6/56,12/50] & [70.33,74.96] & [55.85,61.81] & [20.60,14.60] & [1.71,\textbf{1.29}]\\
ML-RPL \cite{b41} & [6/56,12/50] & [77.95,80.29] & [64.53,67.53] & [\textbf{8.77},9.50] & [2.29,2.21]\\
GIGP (ours) & [6/56,12/50] & [\textbf{79.91},\textbf{81.84}] & [\textbf{66.99},\textbf{69.66}] & [9.21,\textbf{6.19}] & [2.21,1.44]\\
\hline
\end{tabular} 
\label{table2}
\end{threeparttable} 
\end{table}

\begin{table}[!t]
\centering
\caption{Results of quantitative comparison on LA dataset.}
\renewcommand{\arraystretch}{1.2}
\begin{threeparttable} 
\fontsize{8pt}{10pt}\selectfont
\begin{tabular}{@{}cccccc@{}}
\hline
\multirow{2}{*}{Method} & Scans used & \multicolumn{4}{c}{Metrics}\\
\cline{2-6}
& Labeled/Unlabeled & Dice($\%$) $\uparrow$ & Jaccard($\%$) $\uparrow$ & 95HD $\downarrow$ & ASD $\downarrow$\\
\hline
V-Net \cite{b42} & [4/0,8/0] & [52.55,78.57] & [39.60,66.96] & [47.05,21.20] & [9.87,6.07]\\
\hline
DTC \cite{b3} & [4/76,8/72] & [82.75,87.43] & [71.55,78.06] & [13.77,8.38] & [3.91,2.40]\\ 
SASSnet \cite{b4} & [4/76,8/72] & [83.26,87.80] & [71.92,76.91] & [15.51,14.57] & [4.63,4.11]\\
UA-MT \cite{b38} & [4/76,8/72] & [81.16,84.48] & [68.97,73.98] & [24.22,17.13] & [6.97,4.82]\\
URPC \cite{b6} & [4/76,8/72] & [83.47,87.14] & [72.56,77.41] & [14.02,11.79] & [3.68,2.43]\\
MC-Net \cite{b39} & [4/76,8/72] & [84.06,88.99] & [73.04,80.32] & [12.16,7.92] & [2.42,\textbf{1.76}]\\
MC-Net+ \cite{b40} & [4/76,8/72] & [83.13,88.33] & [71.58,79.32] & [12.69,9.07] & [2.71,1.82]\\
AC-MT \cite{b5} & [4/76,8/72] & [87.42,88.74] & [77.83,79.94] & [9.09,8.29] & [2.19,1.91]\\
CAML \cite{b7} & [4/76,8/72] & [87.54,89.44] & [77.95,81.01] & [10.76,10.10] & [2.58,2.09]\\
ML-RPL \cite{b41} & [4/76,8/72] & [84.70,87.35] & [73.75,77.72] & [13.73,8.99] & [3.53,2.17]\\
GIGP (ours) & [4/76,8/72] & [\textbf{89.66},\textbf{90.94}] & [\textbf{81.35},\textbf{83.44}] & [\textbf{5.50},\textbf{5.29}] & [\textbf{1.93},1.81]\\
\hline
\end{tabular} 
\label{table3}
\end{threeparttable} 
\end{table}

\noindent \textbf{Implementation details.} We implement our GIGP using PyTorch 2.4.0 and CUDA 11.8 on an NVIDIA GeForce RTX 4090 GPU. During training, all models were optimized using Stochastic Gradient Descent (SGD) with a weight decay of $3 \times 10^{-5}$, momentum of 0.9, and a learning rate of 0.01. The training process consisted of 300 epochs, with early stopping implemented.  
We employed data augmentation techniques such as random cropping, flipping, rotation, and scaling to reduce overfitting. V-Net \cite{b42} was used as the backbone network. For pancreas segmentation, patches were cropped randomly into $96\times{96}\times{96}$ voxels, with sliding window predictions using strides of $16\times{16}\times{16}$ and $64\times{64}\times{64}$ voxels, respectively. For LA segmentation, input blocks were cropped to $112\times{112}\times{80}$ voxels, with a sliding window stride of $18\times{18}\times{4}$ voxels applied for prediction.

\begin{figure}[!t]
\centering
\includegraphics[width=\textwidth]{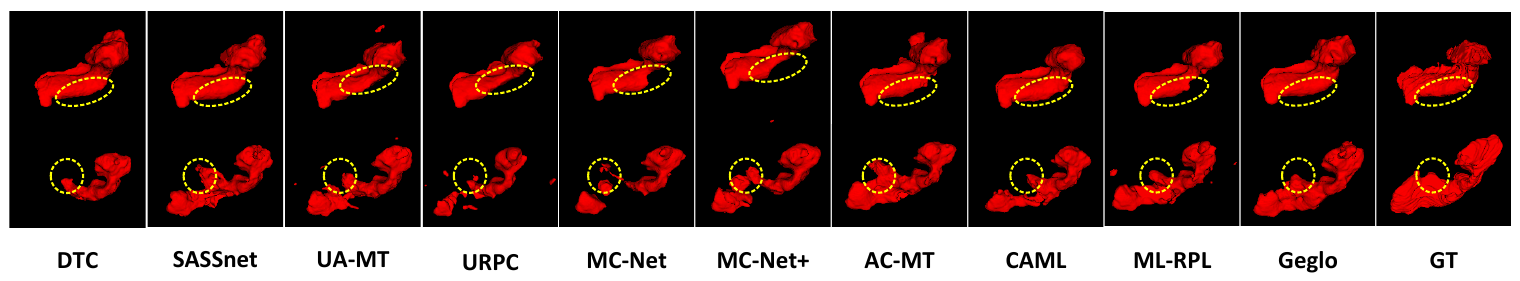}
\caption{Results of qualitative comparison on Pancreas dataset under 20$\%$ labeled data setting. GT represents the ground truth.}
\label{Pancreas}
\end{figure}

\begin{figure}[!t]
\center
\includegraphics[width=\textwidth]{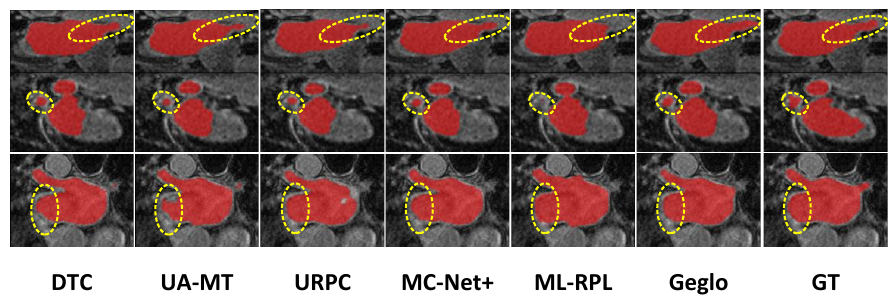} 
\caption{Results of qualitative comparison on LA dataset under 10$\%$ labeled data setting. GT represents the ground truth.}
\label{LA}
\end{figure}

\noindent\textbf{Quantitative Evaluation and Comparison.} As shown in Table \ref{table2} and Table \ref{table3}, our method achieves optimal performance in Dice and Jaccard metrics. It outperforms ML-RPL by 1.96\% and 2.46\%, and by 1.55\% and 2.13\% in Dice and Jaccard scores under 10\% and 20\% labeled data settings on the Pancreas dataset, respectively. GIGP surpasses CAML by 2.12\% and 3.40\%, and by 1.50\% and 2.43\% in Dice and Jaccard scores under 5\% and 10\% labeled data settings on the LA dataset, respectively. These metrics primarily emphasize global structure alignment, demonstrating that focus on learning global geometric information effectively of GIGP aids medical image segmentation. However, ASD and 95HD metrics are not always optimal, as they prioritize local boundary accuracy: ASD measures the average distance between segmentation and ground truth boundaries, while 95HD measures the maximum distance for the top 95$\%$ of points. Thus, there remains room to improve local boundary alignment accuracy.

As shown in Fig. \ref{Pancreas} and Fig. \ref{LA}, GIGP is more complete and closer to the actual facts in terms of both overall performance and details, especially with a noticeable contrast at the yellow dashed line compared with other methods. Our GIGP better preserves the overall features of the data. 

\begin{table}[!t]
\centering
\caption{Ablation studies on Pancreas dataset under 20$\%$ labeled data setting.}
\renewcommand{\arraystretch}{1.2}
\begin{threeparttable} 
\fontsize{8pt}{10pt}\selectfont
\begin{tabular}{@{}cccccccc@{}}
\hline
\multicolumn{4}{c}{Methods} & \multicolumn{4}{c}{Metrics}\\
\hline
MT & GMAM & GGPC & GIIM & Dice($\%$) & Jaccard($\%$) & 95HD & ASD \\
\hline
$\checkmark$ & & &  & 74.43 & 60.53 & 14.93 & 4.61  \\ 
$\checkmark$ & $\checkmark$ & & & 79.86 & 67.02 & 10.41 & 2.18 \\
$\checkmark$ &  & $\checkmark$ & & 80.56 & 68.09 & 8.30 & 1.71 \\
$\checkmark$ & & & $\checkmark$ & 79.45 & 66.78 & 11.00 & 2.83 \\
$\checkmark$ & $\checkmark$ & $\checkmark$ & $\checkmark$ & \textbf{81.84} & \textbf{69.66} & \textbf{6.19} & \textbf{1.44} \\
\hline
\end{tabular} 
\end{threeparttable}
\label{table6}
\end{table}

\noindent\textbf{Ablation study.} We evaluate the effectiveness of the proposed GMAM, GGPC, and GIIM modules using the MT network as the baseline. Table \ref{table6} shows the ablation study results on the pancreas dataset. While the MT network provides a solid foundation for segmentation accuracy, there is significant room for improvement in overall performance and edge details. Integrating GMAM, GGPC, and GIIM modules notably enhances segmentation accuracy and overlap (Dice and Jaccard) while reducing 95HD and ASD values, indicating better boundary approximation and smoothness. Adding any single module to the MT network improves all four metrics, demonstrating the effectiveness of GIGP and validating the contribution of each module.

\section{Conclusion}
\label{section 4}

In this paper, we propose a semi-supervised medical image segmentation framework designed from a global perspective. Firstly, GIIM aligns features between labeled and unlabeled data at corresponding spatial locations to minimize distribution discrepancies. Secondly, GMAM enforces multi-view and multi-scale constraints using geometric moments, offering robust guidance on geometric features. Thirdly, GGPC applies periodic sine wave distortions to simulate organ dynamics and geometric variations, enhancing the ability of the model to learn generalized features. Our method achieves state-of-the-art results on the NIH pancreas and left atrium datasets, validating its effectiveness.

\section{Acknowledgments} 
This work was supported by National Natural Science Foundation of China Grants (No.12101426 and No.12426308), Beijing Outstanding Young Scientist Program (No. JWZQ20240101027), and Beijing Natural Science Foundation Grants (No.Z210003, No.4254093 and L242127).


\begin{thebibliography}{00}

\bibitem{b3} X. Luo, J. Chen, T. Song, and G. Wang, “Semi-supervised medical image
segmentation through dual-task consistency,” in \textit{Proc. AAAI Conf. Artif. Intell.}, 2021, pp. 8801–8809. 

\bibitem{b4}  S. Li, C. Zhang, and X. He, “Shape-aware semi-supervised 3D semantic
segmentation for medical images,” in \textit{Proc. Int. Conf. Med. Image Comput. Comput.- Assist. Intervent.}, 2020, pp. 552–561.

\bibitem{b5} Z. Xu \textit{et al.}, “Ambiguity-selective consistency regularization for mean-teacher semi-supervised medical image segmentation,” \textit{Med. Image Anal.}, vol. 88, pp. 102880, 2023.

\bibitem{b6} X. Luo \textit{et al.}, “Semi-supervised medical image segmentation via uncertainty rectified pyramid consistency,” \textit{Med. Image Anal.}, vol. 80, pp. 102517, 2022.

\bibitem{b7} S. Gao, Z. Zhang, J. Ma, Z. Li, and S. Zhang, “Correlation-aware mutual learning for semi-supervised medical image segmentation,” in \textit{Proc. Int. Conf. Med. Image Comput. Comput.- Assist. Intervent.}, Oct. 2023, pp. 98-108.

\bibitem{b8} X. Guo \textit{et al.} Semi-Supervised Medical Image Segmentation Based on Local Consistency and Volume Constancy. Presented at IEEE Int. Symp. Biomed. Imaging (ISBI). 2024. pp. 1-5.

\bibitem{b52} Y. Xia \textit{et al.}, “Uncertainty-aware multi-view co-training for semi-supervised medical image segmentation and domain adaptation,” \textit{Med. Image Anal.}, vol. 65, no. 101766, 2020.

\bibitem{b57} T. Lei, D. Zhang, X. Du, X. Wang, Y. Wan and A. K. Nandi, “Semi-Supervised Medical Image Segmentation Using Adversarial Consistency Learning and Dynamic Convolution Network,” \textit{IEEE Trans. Med. Imag.}, vol. 42, no. 5, pp. 1265-1277, May. 2023. 

\bibitem{b58} C. Chen, K. Zhou, Z. Wang, and R. Xiao, “Generative consistency for semi-supervised cerebrovascular segmentation from TOF-MRA,” \textit{IEEE Trans. Med. Imaging}, vol. 42, no. 2, pp. 346-353, 2022. 

\bibitem{b20} A. Santhirasekaram, M. Winkler, A. Rockall, and B. Glocker, "A geometric approach to robust medical image segmentation," \textit{Med. Image Anal.}, vol. 97, pp. 103260, 2024.

\bibitem{b21} K. Wang, X. Zhang, Y. Lu, W. Zhang, S. Huang, and D. Yang, "GSAL: Geometric structure adversarial learning for robust medical image segmentation," \textit{Pattern Recognit.}, vol. 140, pp. 109596, 2023.

\bibitem{b22} Y. Meng, \textit{et al.}, "Multi-granularity learning of explicit geometric constraint and contrast for label-efficient medical image segmentation and differentiable clinical function assessment," \textit{Med. Image Anal.}, vol. 95, pp. 103183, 2024.

\bibitem{b60} Bai, Y., Chen, D., Li, Q., Shen, W., and Wang, Y. (2023). Bidirectional copy-paste for semi-supervised medical image segmentation. In Proceedings of the IEEE/CVF conference on computer vision and pattern recognition (pp. 11514-11524).

\bibitem{b13} C.M. Seibold, S. Reiß, J. Kleesiek, and R. Stiefelhagen, “Reference-guided pseudo-label generation for medical semantic segmentation,” in \textit{Proc. AAAI Conf. Artif. Intell.}, 2022, pp. 2171–2179. 

\bibitem{b14} A. Iscen, G. Tolias, Y. Avrithis, and O. Chum, “Label propagation for deep semi-supervised learning,” in \textit{Proc. IEEE/CVF Conf. Comput. Vis. Pattern Recognit.}, 2019, pp. 5070-5079. 

\bibitem{b16} H. R. Roth \textit{et al.}, “Deeporgan: Multi-level deep convolutional networks for automated pancreas segmentation,” in \textit{Proc. Int. Conf. Med. Image Comput. Comput.- Assist. Intervent.}, Munich, Germany, 2015, pp. 556-564. 

\bibitem{b17} Z. Xiong \textit{et al.}, “A global benchmark of algorithms for segmenting the left atrium from late gadolinium-enhanced cardiac magnetic resonance imaging,” \textit{Med. Image Anal.}, vol. 67, pp. 101832, 2021.

\bibitem{b30} Z. Xing, T. Ye, Y. Yang, G. Liu, and L. Zhu, Segmamba: Long-range sequential modeling mamba for 3d medical image segmentation, arXiv preprint, 2024, arXiv:2401.13560.

\bibitem{b43} Gu, A., Dao, T., 2023a. Mamba: Linear-time sequence modeling with selective state spaces. arXiv preprint arXiv:2312.00752.

\bibitem{b36} P. Behjati, P. Rodriguez, C. Fernández, I. Hupont, A. Mehri, and J. Gonzalez, "Single image super-resolution based on directional variance attention network," \textit{Pattern Recognit.}, vol. 133, pp. 108997, 2023.

\bibitem{b42} F. Milletari, N. Navab, and S.-A. Ahmadi, V-Net: Fully convolutional neural networks for volumetric medical image segmentation. Presented at Proc. 4th Int. Conf. 3D Vis. 2016. pp. 565–571.

\bibitem{b38} L. Yu, S. Wang, X. Li, C.-W. Fu, and P.-A. Heng, “Uncertainty-Aware Self-ensembling Model for Semi-supervised 3D Left Atrium Segmentation,” in \textit{Proc. Int. Conf. Med. Image Comput. Comput.- Assist. Intervent.}, 2019, pp. 605-613. 

\bibitem{b39} Y. Wu, M. Xu, Z. Ge, J. Cai, and L. Zhang, “Semi-supervised left atrium
segmentation with mutual consistency training,” in \textit{Proc. Int. Conf. Med. Image Comput. Comput.- Assist. Intervent.}, 2021, pp. 297–306.

\bibitem{b40} Y. Wu \textit{et al.}, “Mutual consistency learning for semi-supervised medical image segmentation,” \textit{Med. Image Anal.}, vol. 81, pp. 102530, 2022.

\bibitem{b41} J. Su, Z. Luo, S. Lian, D. Lin, and S. Li, S., "Mutual learning with reliable pseudo label for semi-supervised medical image segmentation," \textit{Med. Image Anal.} vol. 94, pp. 103111, 2024.

\end{thebibliography}
\end{document}